\documentclass[journal]{IEEEtran}

\usepackage{graphicx}
\usepackage{caption}
\usepackage{multirow}
\usepackage{mathtools}
\usepackage{amsmath}
\usepackage{algorithm}
\usepackage{algorithmic}
\usepackage{amssymb}
\usepackage{xcolor}
\usepackage{tikz}
\usepackage{pgfplots}
\pgfplotsset{compat=1.7}
\usepackage{tikz-3dplot}
\ifCLASSINFOpdf
\else
\fi
\begin{document}
%
\title{Anomaly Detection in Smart Manufacturing with an Application Focus on Robotic Finishing Systems: A Review}
%
%
%

\author{Tareq Tayeh
        and Abdallah Shami\\
        ECE Department, Western University, London, Canada\\
\{ttayeh, abdallah.shami\}@uwo.ca}

\maketitle

\begin{abstract}
As systems in smart manufacturing become increasingly complex, producing an abundance of data, the potential for production failures becomes increasingly more likely. There arises the need to minimize or eradicate production failures, one of which is by means of anomaly detection. However, with the deployment of anomaly detection systems, there are many aspects to be considered. In this paper, an overview of the components, benefits, challenges, methods, and open problems of anomaly detection in smart manufacturing and robotic finishing systems are discussed.
\end{abstract}


%
\IEEEpeerreviewmaketitle

\section{Introduction}
%
%
%
%

\IEEEPARstart{I}{nternet} of Things (IoT) has boomed in recent years, due to the massive increase in the number of physical objects embedded with sensors, which allowed the creation of larger and more interconnected "smart" networks \cite{hhnfv2019}. IoT is defined as a computing paradigm that allows interrelated devices to transfer data over a network and communicate with each other without requiring human-to-human or human-to-computer interaction \cite{al2015internet}. By the end of 2020, there was 6.6 billion active and connected IoT devices worldwide \cite{jamiemoss_2020}, and it is estimated that the potential economical impact of IoT applications will be up to USD \$11.1 trillion per year by 2025 \cite{mckinseycompany_2015}.

IoT is used extensively in a range of applications, such as healthcare, agriculture, transportation, and retail. In particular, the industrial segment was ranked as the top IoT application area in 2020 with a global share of 22\% \cite{understanding_iot_2021}, where it is often referred to as Industrial IoT (IIoT). IIoT entails interconnected sensors and devices that are networked together within industrial applications, which includes manufacturing. The connected network enables the collection, exchange, and analysis of data, facilitating improvements in industrial efficiency and overall productivity, which in turn provides cost reduction, shorter time-to-market, mass customization, sustainable production, and improved safety \cite{BOYES20181}. Accenture estimates that IIoT could add up to USD \$14.2 trillion to the global economy by 2030 \cite{accenture}.

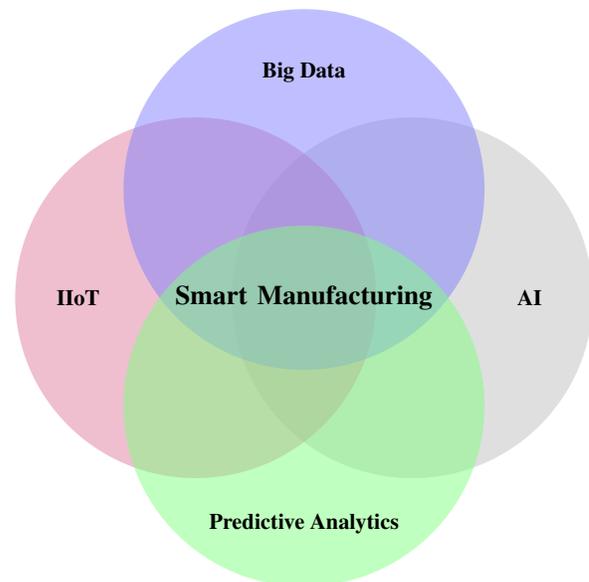
\begin{figure}[!ht]
\centering
\begin{tikzpicture}[thick,scale=0.6, every node/.style={scale=0.6}]
\begin{scope}[opacity=0.5] 
    \fill[gray!50!white]   ( 0:2.4) circle (4);
    \fill[purple!50!white]   ( 180:2.4) circle (4);
     \fill[blue!50!white]   ( 90:2.4) circle (4);
     \fill[green!50!white] (270:2.4) circle (4);
 \end{scope}
  \node at (0:5)  [font=\Large]    {\textbf{AI}};
  \node at (180:5) [font=\Large]   {\textbf{IIoT}};
  \node at (90:5)  [font=\Large]  {\textbf{Big Data}};
  \node at (270:5) [font=\Large]   {\textbf{Predictive Analytics}};
  \node [font=\LARGE] {\textbf{Smart Manufacturing}};
\end{tikzpicture}
\caption{Venn Diagram of Smart Manufacturing}
\label{fig:smartmanufacturingvenn}
\end{figure}

As IIoT systems are becoming increasingly complex and produce diverse data at an ever-increasing rate, the incorporation of computer intelligence is needed to capture the value that IIoT applications generate, resulting in the emergence of smart manufacturing. Smart manufacturing utilizes intelligent computer-controlled and internet-connected machinery to harness sensor data and automate operations to enhance the manufacturing supply chain and performance. In essence, smart manufacturing integrates IIoT with predictive analytics, big data, and Artificial Intelligence (AI), as visualized in Figure \ref{fig:smartmanufacturingvenn}, enabling a more intelligent insight into manufacturing processes.

Statistics show that 82\% of the companies that utilize smart manufacturing technologies have experienced an increase in their manufacturing efficiency and 45\% have experienced an increase in customer satisfaction \cite{ASQ}. However, due to the complex nature of the automated, interconnected systems in smart manufacturing; potential production failures might occur. This raises the need to detect anomalies efficiently across the entire manufacturing process in order for the operators to solve the underlying issues and achieve the required product quality.  In this paper, the background, benefits, challenges, methods, and open problems of anomaly detection in smart manufacturing, with an application focus on robotic finishing systems, are discussed.

\section{Background}
In this section, robotic finishing systems and anomaly detection are introduced and discussed.

\subsection{Robotic Finishing Systems}
All manufactured products undergo at least a single finishing session during their production before they can be used. A finishing session aims to alter the surface of a manufactured part in order to achieve a particular characteristic \cite{3d_experience}. Traditionally, finishing sessions have been conducted manually by human experts, particularly in small and medium volume productions where the finishing tasks are non-repetitive in nature. However, manual finishing can be labor-intensive and impose health and safety risks to the human workers. Manual finishing can add significant costs to the overall manufacturing process, as well as reducing the product manufacturing rate, especially when all other stages of the manufacturing process are automated \cite{pires2009finishing}. As a result, automated mechanisms, and in particular, robotic-based processes, have emerged for finishing tasks instead \cite{pires2009finishing}. 

Robotic finishing systems usually indicate either equipping a standard robotic arm with material removal devices, such as abrasive stones and grinding tools, or using the standard robotic arm to present the product to fixed finishing tools. The systems can be used for light-duty tasks such as buffing, lapping, or polishing, and for heavier operations such as weld bead finishing or burr removal. Furthermore, the introduction of Computer Numeric Control (CNC) machines allow more complex parts to be processed, as it provides the ability for program computer devices to control machine tools, rapidly advancing productivity by automating the highly technical and labor-intensive processes. Moreover, CNC machines allow the operator to call a program from memory, change parameters, and resume production in a matter of seconds or minutes.

Robotic finishing systems include \cite{mesh_automation_2020, fanuc}:

\begin{enumerate}
    \item Grinding Robots: Robots that use an abrasive wheel as the cutting tool for grinding. Each abrasive part on the wheel's surface cuts a small chip off the manufactured product workpiece via shear deformation.
    \item Buffing Robots: Robots that enhance metal surfaces by removing its dull, smudged, or scratched outlook and smoothing it to give a new, refined, shiny or appealing look.
    \item Polishing Robots: Similar to buffing robots, as both procedures include changing the workpiece surface to become more appealing, however, polishing involves the use of abrasive products to even out imperfections.
    \item Sanding Robots: Robots that utilize a moving abrasive surface or sandpaper to smooth and remove workpiece surfaces to prepare it for painting or repairment.
    \item Cutting Robots: Robots that cut or carve the workpiece into a particular shape and size.
    \item Deburring Robots: Robots that remove small imperfections known as burrs from grinded, drilled, stamped, or machined sharp edges, and from corners on machined parts, molded components, forgings, castings, and welded assemblies.
    \item Deflashing Robots: Robots that remove dross, flash, or parting lines on castings, forgings, molded components and welded assemblies via tumbling, breaking, grinding, or cutting. Could be considered as a subset of deburring robots in some aspects.
    \item Routing Robots: Robots that rout an area in hard material via a rotating blade.
    \item Drilling Robots: Robots that cut into a surface vertically by digging holes using up and down motions.
    \item Milling Robots: Robots that cut into a surface vertically and horizontally with the side of the bit, digging holes using up, down, left, and right motions. Similar to drilling robots but with a wider range of use and bulkier machinery. 
\end{enumerate}

Utilizing these various robotic finishing systems enables efficient material removal, reduced waste production, reduced tool wear, longer tool life, increased health and wellbeing, improved energy use, and reduced harmful emissions \cite{Robotiq}. 

Moreover, smart manufacturing provides the required visibility into shop floor and field operations to enable product quality control and anomaly detection, in a process called condition monitoring. Condition monitoring can be conducted by two methods: by inspecting products as it moves through the production cycle, or by monitoring the condition of the machinery and tools which manufacture the products \cite{shiklo_2018}. The former method provides more accurate results by uncovering minor defects, however, it can only be applicable to discrete manufacturing and such inspections are performed manually, making it a costly, laborious and time-sensitive operation. The latter, on the other hand, helps to detect bottlenecks in the manufacturing operation by identifying badly tuned or underperforming machines when defected products are manufactured, making it a more suitable method of use in majority of the cases.

However, even with these advantages and advances in smart manufacturing and in automated robotic finishing systems, it is still not guaranteed that a product will be manufactured free of defects, raising the need for efficient anomaly detection methods.

\subsection{Anomaly Detection}
Anomaly detection, also called novelty detection, outlier detection, or event detection, is the process of identifying unusual or novel observations or sequences within the data captured. These anomalous sequences are often referred to as anomalies, outliers, discordant observations, exceptions, faults, defects, aberrations, noise, errors, damage, surprise, novelty, peculiarities or contaminants in different application domains \cite{chandola2007outlier}. Furthermore, there have been various attempts to define the nature of anomalous data. Hawkins defined an anomaly as "An observation which deviates so significantly from other observations as to arouse suspicion that it was generated by a different mechanism." \cite{hawkins1980identification}. Alternatively, Barnett and Lewis defined an anomaly as "An outlier is an observation (or subset of observations) which appears to be inconsistent with the remainder of that set of data." \cite{barnett1984outliers}.

Anomaly detection is an important concept in the data analysis domain and has been widely researched in a variety of domains such as fraud detection, fault detection, and intrusion detection in diverse application domains such as health care, tax, insurance, finance, cyber security, traffic management, energy management, automated industrial processes, and many others. Anomalous data can indicate significant and critical incidents that can require urgent attention. In smart manufacturing, an anomaly is considered as an unanticipated change in the status or behavior of the IIoT system which deviates from the norm. The nature of IIoT data follows two important observations \cite{cook2019anomaly}:

\begin{enumerate}
    \item The majority of the data captured by an IIoT system is considered non-anomalous, representing the normal operating status of the system.
    \item The normal operating behavior of the system can change at any time for many reasons.
\end{enumerate}

\par \noindent Moreover, there are three broad categories of time series anomalies \cite{cook2019anomaly}:

\begin{enumerate}
    \item Point Anomaly: A single data point that deviates significantly from other observations before the time series returns to its previous normal state. Point anomalies could be the result of statistical noise or faulty sensors. 
    \item Contextual Anomalies: Sequence of observations which deviate from the expected time series patterns in the same context. However, if these observations were taken individually, their values may be within the expected range of values for the time series. Contextual anomalies could be the result of seasonal patterns.
    \item Collective Anomalies: Collection of observation which deviate from the entire data set, but the values of the individual data points are not considered anomalies contextually. Collective anomalies could be the result of combining time series data from two sensors that combined indicate an anomaly.
\end{enumerate}

Many anomaly detection methods require substantial human effort to facilitate a smart manufacturing system, including data interpretation and analysis. It is easier for an expert to manually identify trends and patterns in a small subset of data representing the system, however, as the number of interconnected devices and machines increase, the complexity of the system and data analysis increases. This, in turn, increases the need for developing automated and intelligent anomaly detection approaches to enable experts to focus solely on the most important observations. One of the main purposes of utilizing intelligent anomaly detection approaches is to provide a deeper insight into the operational state of these devices, improving the overall process efficiency. In addition, as the prices of IIoT devices fall, the likelihood of replacing older industrial equipment with new retrofitted monitoring devices increases. 

Financially, it is estimated that a 1\% productivity improvement across the manufacturing industry can result in an annual savings of USD \$500 million, whereas a breakdown in the production line can cost up to USD \$50 thousand per hour \cite{ProgressDataRPM}. Furthermore, predicting anomalies on time can decrease the number of breakdowns by up to 70\%, maintenance costs by up to 30\%, and over-scheduled repairs by up to 12\% \cite{ProgressDataRPM}.

\section{Challenges}

Maximizing machine availability and reducing the number of defected products is challenging. Any abrupt machine failure would result in an undesirable loss of quality and low assembly line uptime would result in loss of productivity \cite{ProgressDataRPM}. Anomaly detection helps ease and reduce these challenges, however, there are a number of factors which make anomaly detection itself very challenging in the smart manufacturing domain. The main factors include \cite{cook2019anomaly}:

\begin{enumerate}

    \item Historical Data: It is challenging to have sufficient historical data to model and define all the correct normal and anomalous states of a system when deploying an anomaly detection approach into a poorly known or novel system \cite{chalapathy2019deep}. Moreover, there is often an abundance of non-anomalous data and a relatively small or no amount of anomalous data in well-optimized processes, making supervised learning-based approaches infeasible. In order to tackle this challenge, a number of methods exist such as manipulating the imbalanced data sets, introducing copies of known anomalies or synthetic anomalies, or reducing non-anomalous instances. However, all these approaches can damage the temporal context between the time series, as important trends may be eliminated. Unsupervised or semi-supervised approaches tackle this lack of knowledge by training the system using only the normal data collected about the system’s state so that when data falls outside of the system state’s boundary it is considered as anomalous. Unsupervised or semi-supervised approaches allow the discovery of novel anomalies or anomaly modes in a new environment, but at a cost of having detailed information about the discovered anomalies.

    \item Data Dimensionality: Refers to the number of separate data attributes represented in each observation \cite{chandola2009kumar}. Some anomaly detection approaches are unsuitable for higher dimensional data, which incur higher computational costs than lower dimensional data. IIoT data is produced in two main categories: \\
    a) Univariate Data: A single sensor producing a sequence of observations or aggregated data from multiple sensors combined into a single sequence of observations is considered as univariate data. Univariate data are often in the form of key-value pair, where the key is the timestamp of the observation and the value is the corresponding sensor's reading. For example, univariate data could just indicate timestamps and the corresponding robotic finishing machine's feedrate. \\
    b) Multivariate Data: Multiple sensors each producing a sequence of observations, which can be viewed as a collection of temporally correlated univariate data that provide a more thorough and rich view of the system, are considered as multivariate data. Multivariate data are often in the form key-vector pair, where the key is the timestamp of the observation and the vector is the readings associated with the different sensors. Furthermore, multivariate data requires to analyze the relationship between the various univariate data at given time steps to enrich the anomaly detection algorithm. For example, multivariate data could indicate timestamps and the corresponding readings from the output current, position, and velocity in a robotic finishing machine.
    
    \item Stationarity: A stationary time series is one where the mean, variance, and autocorrelation does not depend on the time at which the series is observed. A system needs to display stationarity in order to effectively utilize many smart manufacturing anomaly detection approaches. Furthermore, it is important for anomaly detection approaches to adapt to the changes in data structures for longer-term deployments, as historic anomalous data points might not be anomalous anymore given the current state of the system. Non-stationary elements include: \\
    a) Seasonality: Cyclic patterns that repeat regularly over time. \\
    b) Trends: The linear increasing or decreasing behavior of the series over time. \\
    c) Concept Drift: Change in the statistical properties of the data flow over time in unforeseen ways. \cite{gama2014survey}. \\
    d) Change Points: Local or global permanent changes in the normal state of the system \cite{basseville1993detection}. These changes are more sudden and rapid than those in concept drift, and the system rapidly adapts to a new state. Change points can be expected when machine or machine component upgrades occur, or an unexpected increase in usage of a particular component.
    
    \item Noise: Represents fluctuations and unwanted random disturbance in the data caused by unrelated events within the sensor’s surroundings, transmission errors in the data management system, or by slight differences in the sensitivity of the detector. Although noisy data do not affect the overall structure of the data, it is important to understand its nature and the reasons behind it as it may represent a major event within the system that requires the use of complex noise reduction techniques.

    \item Contextual Information: The abundance of data available enables the inclusion of contextual information into the anomaly detection method \cite{hayes2014contextual}. The inclusion of contextual information enriches the anomaly detection algorithm to correctly identify anomalous observations or sequences that do not conform to the norm, but increases the complexity of the process and introduces some challenges that must be overcome, such as: \\
    a) Temporal Context: It is implied that a temporal correlation between observations exist in the generated IIoT time series data \cite{sezer2017context}. For example, there exists temporal correlation between the output current, position, and velocity in a robotic finishing machine. \\
    b) Spatial Context: It is implied that a spatial context exists when there are multiple sensors deployed monitoring a system, such as sensor's location and activity, time of day, and sensor's proximity to other devices \cite{sezer2017context}. Spatial context becomes more challenging as it increases in size or when sensors are made mobile. For example, if sensors on the robotic finishing system were placed at different elevations or angles, observations that were previously normal at specific time steps may be anomalous when observed as the robotic finishing system conducts its task. \\
    c) External Context: A subset of spatial context that monitors the external conditions around the system. For example, external context could be monitoring the temperature and humidity of the manufacturing facility in which the robotic finishing system is placed in, by mounting external sensors on the system.
    
    \item Time and Resource Constraints: The majority of IIoT devices have limited computational resources, making it challenging to process any data on these devices. Therefore recently, models are being trained and processed at a centralized location, usually in the cloud or at a datacenter, where higher computational resources are being leveraged. However, centralized processing introduces latency due to the round-trip delays and resource scheduling \cite{bose2019adepos}. In many cases, having a bit of latency does not raise issues, however, in some cases, it is a requirement to process the data rapidly and be able to generate reports as soon as the data is generated \cite{razzaque2015middleware}. The use of Fog and Edge devices aim to perform the data processing closer to the location of the data, however, these devices have lower power than cloud services or datacenters, highlighting the importance of realizing the computational cost of any operation performed on the data during deployment \cite{razzaque2015middleware}. Moreover, managing and storing data coming from multiple sensors can cause a concern due to its volume and velocity. More specifically, storing an entire data set collected by an IIoT network in a format easily interpreted or retrieved by the anomaly detection approach may be impractical. As a workaround, analysis and computations can be performed before further data is collected in techniques such as: \\
    a) Sliding Windows: Indicates creating a specified window size and performing calculations on the data within this window, before sliding or rolling to the next window based on the step size specified and so on, till the entire data has been covered in at least a single window. Sliding windows help reduce the storage requirements on devices, however, some attributes and features may not be discovered when analysis is only conducted on particular windows. Hence, anomaly detection models would need a way of retaining past trends and patterns without having access to the entire data history. \\
    b) Incremental Processing: Indicates processing the most recent observation only, where each data point is analyzed exactly once by the anomaly detection model. Using incremental processing requires all past trends and patterns to be retained within the model.
    
    \item Reporting Anomalies: There are two main ways to report anomalies \cite{goldstein2016comparative}: \\
    a) Anomaly Score: Usually a value between 0 and 100 that indicates the degree of deviation of a data point from the expected value, where a high deviation would translate to a high anomaly score. This method is useful when anomalies are to be analyzed further or when investigating certain types of anomalies within a particular time period. For example, the use of anomaly score would be ideal when different types of anomalies exist in robotic finishing machines and these anomalies need to be analyzed further.
    b) Labels: A binary label indicating whether a data point observation is anomalous or non-anomalous. The assigned label is often calculated by setting a threshold and any value that crosses the set threshold is labeled as anomalous. This method is useful when immediate reporting is required or when having binary labels is sufficient for the efficiency of the anomaly detection approach. For example, the use of labels would be ideal when the finished component part is to be visually inspected for any defects.
    
\end{enumerate}

The discussed factors and their challenges are also summarized in Table \ref{tab:adsmsummary}.

\begin{table}[htb]
\centering
\begin{tabular}{|l|l|}
\hline
\textbf{Factor}                                                          & \textbf{Challenge}                                                                                                                                                                                                                                              \\ \hline
Historical Data                                                          & \begin{tabular}[c]{@{}l@{}}- Challenging to have sufficient historical data\\ to model all the correct normal and anomalous\\ states of a system.\\ - Challenging to obtain anomalous data in \\ well-optimized processes.\end{tabular}                          \\ \hline
Data Dimensionality                                                      & \begin{tabular}[c]{@{}l@{}}- Challenging to deal with high dimensonal \\ data, where many sensors are each producing \\ a sequence of observations.\end{tabular}                                                                                                   \\ \hline
Stationarity                                                             & \begin{tabular}[c]{@{}l@{}}- Challenging to deal with a time series \\ where its mean, variance, and autocorrelation \\ are time-dependent.\end{tabular}                                                                                                           \\ \hline
Noise                                                                    & \begin{tabular}[c]{@{}l@{}}- Challenging to deal with major fluctuations \\ and unwanted random disturbances in the data.\end{tabular}                                                                                                                          \\ \hline
Contextual Information                                                   & \begin{tabular}[c]{@{}l@{}}- Challenging to deal with temporal, spatial, \\ and external contextual information to enrich \\ anomaly detectors.\end{tabular}                                                                                                       \\ \hline
\begin{tabular}[c]{@{}l@{}}Time and Resource \\ Constraints\end{tabular} & \begin{tabular}[c]{@{}l@{}}- Challenging to process data on IIoT devices \\ due to their limited computational resources.\\ - Challenging to store the entire data produced \\ in a practical format for the anomaly detectors.\end{tabular}                    \\ \hline
Reporting Anomalies                                                      & \begin{tabular}[c]{@{}l@{}}- Challenging to calculate the correct anomaly \\ score for each obtained observation during \\ deployment.\\ - Challenging to set the most appropriate \\ anomaly threshold for observations collected \\ during deployment.\end{tabular} \\ \hline
\end{tabular}
\caption{Summary of Anomaly Detection Challenges in Smart Manufacturing}
\label{tab:adsmsummary}
\end{table}

\section{Methods}

Effective anomaly detection in robotic finishing systems can be split into two main domains: time series analysis and computer vision. Firstly, to monitor and analyze robotic machine components via sensor readings to mitigate anomalous outcomes, time series analysis-based anomaly detection methods were developed. Secondly, to analyze the visual quality of finished products and mitigate finishing defects, automated computer vision-based anomaly detection methods were developed. In this section, each domain and its general methods will be discussed in detail.

\subsection{Time Series Analysis} \label{timeseriesanalysis}

Time series is defined as an ordered sequence of data points at equally spaced time intervals. Time series analysis is a statistical technique for analyzing this time series data, which enables time series models to be built to obtain an understanding of the underlying nature of the time series data or fit a model for forecasting or monitoring. An important application of time series analysis and modelling is anomaly detection, where novel or unusual states within a system are identified.

To examine the quality of the manufacturing process, components such as equipment calibration, machine conditions, and environmental conditions are monitored to ensure that their status are within their normal range and identify when they exceed beyond the normal thresholds. Ideally, if sensor readings are approaching thresholds that can lead to potential product defects, a monitoring solution should trigger an alert and pinpoint the source of the problem in order for the operators to be able to solve the underlying problems.

The general methods for detecting anomalies in time series data can be divided into the following seven groups:

\begin{enumerate}
    \item Statistical and Probabilistic: Methods that utilize historical data to model the normal behavior of the system, where if a new observation received by the system does not fit the model, the observation is labelled as an anomaly \cite{markou2003novelty, saci2021autocorrelationn}.
    \item Predictive: Methods that generate a regression model based on historic system trends, where if a new observation received by the system vary greatly from the predicted values, the observation is labelled as an anomaly \cite{giannoni2018anomaly}.
    \item Reconstructive: Methods that embed data into a lower dimensional subspace and then reconstructs them, where if a new observation received by the system vary greatly from the expected values, the observation is labelled as an anomaly.
    \item Pattern Matching: Methods that use direct modeling of the time series, where if a new observation received by the system follows known characteristics of anomalous subsequences based on a database of labelled anomalies or historic patterns within normal data, the observation is labelled as an anomaly.
    \item Distance-Based: Methods that utilize distance metrics to model relationships between observations, where if a new observation possess a large enough distance of separation with the normal data, the observation is labelled as an anomaly \cite{chandola2009kumar}.
    \item Clustering: Methods that project the data into a multi-dimensional space to create clusters, where if a new observation received by the system is further away or does not belong to the dense clusters, the observation is labelled as an anomaly \cite{chandola2009kumar}.
    \item Ensemble: Methods that utilize different algorithms to observe data points, where a voting mechanism is employed over the results of each algorithm to produce the resulting ensemble method output. \cite{yangIoTJournal20211}
\end{enumerate}

Table \ref{tab:prosconsad} discusses pros and cons of the aforementioned general methods.

\begin{table}[h]
\centering
\begin{tabular}{|l|l|l|}
\hline
\textbf{Method}                                                          & \textbf{Pros}                                                                                                                               & \textbf{Cons}                                                                                                              \\ \hline
\begin{tabular}[c]{@{}l@{}}Statistical and \\ Probabilistic\end{tabular} & \begin{tabular}[c]{@{}l@{}}Often easier to detect \\ outliers as real-life data \\ often follows some \\ statistical distribution.\end{tabular} & \begin{tabular}[c]{@{}l@{}}Often challenging with \\ high dimensional data.\end{tabular}                                   \\ \hline
Predictive                                                               & \begin{tabular}[c]{@{}l@{}}Useful for real-time \\ prediction of anomalies.\end{tabular}                                                     & \begin{tabular}[c]{@{}l@{}}Often requires large \\ volume of training data.\end{tabular}                                    \\ \hline
Reconstructive                                                           & \begin{tabular}[c]{@{}l@{}}Able to detect novel \\ anomalies.\end{tabular}                                                                  & \begin{tabular}[c]{@{}l@{}}Often requires large \\ volume of training data.\end{tabular}                                    \\ \hline
Pattern Matching                                                         & \begin{tabular}[c]{@{}l@{}}Robust in detecting \\ anomalies.\end{tabular}                                                          & \begin{tabular}[c]{@{}l@{}}Often requires large \\ volume of training data.\end{tabular}                                    \\ \hline
Distance-Based                                                           & \begin{tabular}[c]{@{}l@{}}Works well when \\ normal data is highly \\ concentrated around a \\ region.\end{tabular}                            & \begin{tabular}[c]{@{}l@{}}Highly dependent on the \\ distance measures and \\ anomaly scoring \\ mechanisms used.\end{tabular} \\ \hline
Clustering                                                               & Unsupervised method.                                                                                                                        & \begin{tabular}[c]{@{}l@{}}Can mistake data points \\ in less dense clusters \\ as anomalies.\end{tabular}                    \\ \hline
Ensemble                                                                 & \begin{tabular}[c]{@{}l@{}}Often improves the \\ anomaly detection \\ performance.\end{tabular}                                                & \begin{tabular}[c]{@{}l@{}}Increases computation \\ time and complexity.\end{tabular}                                      \\ \hline
\end{tabular}
\caption{Pros and Cons of the General Anomaly Detection Methods}
\label{tab:prosconsad}
\end{table}

\subsection{Computer Vision}

Computer vision is a field concerned with enabling computers to gain high-level understanding from digital images or videos, by identifying and processing images in the same way that human vision does. This interdisciplinary field aims to simulate and automate elements of human vision by using sensors and AI algorithms, emulating human intelligence and instincts in a computer. Anomaly detection is one of the most important applications of computer vision, where novel or unusual images or parts of an image within a system are identified.  

Moreover, with the increasing demand for maintaining an almost perfect product quality finish in smart manufacturing, surface finishes are being monitored by cameras installed on various machines to ensure the surface finish roughness and appearance are within the acceptable normal visual bounds. Ideally, if a surface finish does not meet the required product standard and falls out of the acceptable normal bounds, then the manufacturing process will be flagged and will request the intervention of system operators for necessary further examination.

There are four primary computer vision techniques for detecting anomalies in images:

\begin{enumerate}
    \item Image Classification: Technique that aims to comprehend an image as a whole to classify the entire image as normal or anomalous.
    \item Object Detection: Technique that aims to comprehend an entire image to locate and classify potential anomalous objects within an image.
    \item Semantic Segmentation: Technique that aims to segment an entire image and label each pixel to detect potential anomalous objects within an image. Treats multiple objects of the same class as a single entity.
    \item Instance Segmentation: Technique that aims to segment an entire image and label each pixel to detect potential anomalous objects within an image. Treats multiple objects of the same class as distinct individual objects.
\end{enumerate}

Moreover, the general methods mentioned in \ref{timeseriesanalysis} can also be classified as the general methods for anomaly detection in the computer vision domain, but they aim to achieve one of the aforementioned four primary computer vision techniques for anomaly detection \cite{santhosh2020anomaly}.


\section{Open Problems}

Although anomaly detection methods are becoming increasingly accurate and efficient, there are still research challenges that require further exploration. The following are some of the open problems faced by anomaly detection methods in smart manufacturing \cite{cook2019anomaly, tayeh2020distancee, pang2020deep, yangIoTMag20211, aburakhia2020transferrr}:

\begin{enumerate}
    \item Real-Time Processing: In smart manufacturing, detecting anomalies in real-time or near real-time is of utmost importance due to the nature of the manufacturing process. Defective products and machine components need to be detected instantly in order for the operators to fix the underlying issues. Otherwise, if the anomaly detector takes a long time to process observations to flag anomalies, the system will fail, imposing financial and operational problems. 
    \item Reducing Storage Requirements: It would be very costly to hold all the data generated in smart manufacturing for processing and detecting anomalies. Techniques such as sliding windows and incremental processing help reduce the storage and memory requirements for the processing system, but more novel methods could be explored in the future to allow a reduction of storage requirements while processing a larger number of observations.
    \item Change in Data Structures: As data may change during longer-term deployments, an anomaly detection method should be able to identify this data change and still be able to efficiently detect anomalies. However, completely retraining the method with the new data structures is costly, raising the need for online adaptive learning methods to be developed, in order to adapt to novel behavior in the data. Offline methods could be used prior to the the deployment of the online adaptive method for initial method training on the historical data collected from the system.
    \item Insufficient Labeled Anomaly Data: In the real-world, manufacturing processing are often well-optimized, meaning that there will be an abundance of non-anomalous data but a relatively small amount of anomalous data. The lack of anomalous data could impose a limitation on supervised learning approaches, due to the data imbalance and that these scarce anomalous data will often not represent the whole range of potential anomalies that could occur in the system. As a result, efficient unsupervised and semi-supervised approaches need to be developed to ensure the anomaly detection methods are able to detect novel anomalies during deployment.
    \item Utilizing More Data Sources: Utilizing different types of data from various data sources, such as the inclusion of contextual information or incorporating more sensors around the manufacturing systems, can enable the anomaly detection algorithm to learn enriched data behaviors and structures that can assist with detecting hidden anomalies. Therefore, it is important for anomaly detection algorithms to be able to process multivariate data successfully.
    \item Developing a Generalized Anomaly Detection Method: The development of generalized anomaly detection algorithms than can be applied in multiple different areas in smart manufacturing or different domains enables the reusability of methods and eases their deployments in new environments. 
    \item Interpretability of Detected Anomalies: Many anomaly detection methods are proficient at detecting anomalies, however, most also fail in providing clear and concise reasons behind the anomaly detection, particularly in rare anomalies which can lead to algorithmic bias against particular sensors or machines. To mitigate this risk, anomaly detection algorithms should be able to provide straightforward cues about why certain data points are flagged as anomalous, 
    enabling human experts to understand the system better and correct the bias if needed.
\end{enumerate}

\section{Conclusion}

Deploying anomaly detection algorithms in smart manufacturing reduces many financial and operational risks, while reducing the monitoring burden on system operators. However, these algorithms are associated with many challenges which need to be taken into account. These challenges are dictated by the system's specific application, objective, and environment, and need to be addressed when designing and developing the algorithms. Computer vision methods and time series analysis methods are the two most common domain of anomaly detection methods in smart manufacturing, where the former aims to detect anomalies in sensor data and the latter aims to detect anomalies in the produced surface finishes. Although anomaly detection methods are constantly improving with time, many open problems still exist, meaning anomaly detection algorithms still have a significant amount of room to evolve and become more efficient, accurate, and generalized.


\ifCLASSOPTIONcaptionsoff
  \newpage
\fi




\bibliographystyle{ieeetr}
\bibliography{references.bib}

\end{document}